\documentclass[11pt,a4paper]{article}
\usepackage[hyperref]{acl2021}
\usepackage{times}
\usepackage{latexsym}

\usepackage{microtype}
\usepackage{multirow}
\usepackage{graphicx}
\graphicspath{ {images/} }

\aclfinalcopy % Uncomment this line for the final submission
\title{Multilingual Neural Semantic Parsing for Low-Resourced Languages}

\author{Menglin Xia \\
	Amazon Research Cambridge \\
	\texttt{ximengli@amazon.com} \\\And
	Emilio Monti \\
	Amazon Research Cambridge \\
	\texttt{monti@amazon.com} \\}

\date{}

\begin{document}
\maketitle
\begin{abstract}
Multilingual semantic parsing is a cost-effective method that allows a single model to understand different languages. However, researchers face a great imbalance of availability of training data, with English being resource rich, and other languages having much less data. To tackle the data limitation problem, we propose using machine translation to bootstrap multilingual training data from the more abundant English data. To compensate for the data quality of machine translated training data, we utilize transfer learning from pretrained multilingual encoders to further improve the model. To evaluate our multilingual models on human-written sentences as opposed to machine translated ones, we introduce a new multilingual semantic parsing dataset in English, Italian and Japanese based on the Facebook Task Oriented Parsing (TOP) dataset. We show that joint multilingual training with pretrained encoders substantially outperforms our baselines on the TOP dataset and outperforms the state-of-the-art model on the public NLMaps dataset. We also establish a new baseline for zero-shot learning on the TOP dataset. We find that a semantic parser trained only on English data achieves a zero-shot performance of 44.9\% exact-match accuracy on Italian sentences. 
\end{abstract}

\section{Introduction}

Semantic parsing is defined as the task of parsing a natural language sentence into a logical form that represents its meaning. The logical form, or sometimes called the meaning representation language (MRL) expression, can be executed against a knowledge base to extract information; therefore, semantic parsing often finds its application in question answering, code generation, information retrieval, etc. Due to its wide range of applications, semantic parsing has drawn a lot of research interest. Among them, neural semantic parsing methods have gained popularity in recent years due to their good results \cite{dong-lapata-2018-coarse}. Neural semantic parsing often formulates the task as a machine translation problem and uses neural networks to translate the sentences into MRL expressions.

Multilingual neural semantic parsing is a cost-effective method that allows a single model to understand different languages. However, similar to other machine-learning based approaches, neural semantic parsing requires large amounts of training data. To understand texts in different languages, semantic parsing models need training data for each target language. Unfortunately, researchers face a great imbalance of availability of training data for semantic parsing: while we have lots of data in English, the data in non-English languages is often scarce. Although there is a growing number of datasets published for semantic parsing in English, very few datasets are available in other languages. Moreover, manually annotating data for semantic parsing is difficult and time-consuming, as it requires a lot of training and effort for annotators to write MRLs. 

Instead of manually annotating semantic parsing data in low-resourced languages, can we bootstrap training data for multilingual semantic parsing from the more abundant English data? In this paper, we aim to tackle the data limitation problem for multilingual semantic parsing with machine translation. We machine translate English sentences into target non-English languages and make use of the alignment information in the English MRL to create MRL annotations in other languages (see Section \ref{sec:data}). We then describe our methods to build multilingual semantic parsing models on the machine translated training data (see Section \ref{models}). To train the multilingual semantic parser, we mix the training data from all languages together and train a model from scratch (see Section \ref{sec:baseline}). We base our neural semantic parser on the sequence-to-sequence model with pointer mechanism \cite{sutskever2014sequence, NIPS2015_5866}, where both the natural language question and the target MRL are treated as sequences of tokens and the parser learns from the training data a mapping to translate questions into MRLs.  

The machine translation-based data generation method allows us to easily extend English data to other languages. However, the quality of the bootstrapped training data is constrained by the accuracy of the machine translation model and other components of the generation method, such as alignment. To mitigate the problem of data quality of the machine translated training data, we make use of transfer learning with pretrained multilingual encoders to further improve the multilingual semantic parsing model (see Section \ref{sec:pretrain}). 

To evaluate the model performance on sentences written by human as opposed to machine translated ones, we introduce a new multilingual semantic parsing dataset based on the Facebook Task Oriented Parsing (TOP) dataset \cite{top-paper}. We compare our method against several baselines, including monolingual models and a popular technique in literature that relies on translating the utterances and using an English model to understand them (see Section \ref{sec:other_baselines}). We report the experimental results and our analysis in Section \ref{sec:res}. To show that our multilingual semantic parsing models also work with human-generated training data and to compare them against previous work, we report the performance of our models on the public multilingual NLMaps dataset in Section \ref{sec:nlmaps}. 

Apart from bootstrapping training data, zero-shot learning is also a technique that allows a multilingual model to generalize to low-resourced languages. We study how the multilingual semantic parsers with pretrained encoders can generalize to other languages in a zero-shot scenario (see Section \ref{sec:zero_shot}).

Our main contributions are as follows:
\begin{enumerate}
	\item We propose a method to automatically generate training data for multilingual semantic parsing from existing English data via machine translation and we use pretrained multilingual encoders to compensate for the data quality. We release a new multilingual semantic parsing dataset in English, Italian and Japanese based on the public TOP dataset, with \textasciitilde30k machine-translated training and validation data and \textasciitilde8k manually translated test data for each language.
	
	The dataset is available for download at: \url{https://github.com/awslabs/multilingual-top}.
	\item We show that our multilingual semantic parsing model achieves state-of-the-art performance, outperforming several baselines on the TOP dataset and existing work on the public NLMaps dataset. 
	\item We establish a new baseline for zero-shot learning on the TOP dataset with semantic parsing model finetuned from pretrained multilingual encoders. 
\end{enumerate}

\section{Background and Related Work}

Semantic parsing has been studied for a few decades. Earlier methods on semantic parsing rely on defining semantic rules to parse the input sentence \cite{zelle1996, zettlemoyer2005}. With recent advances in neural networks, there is a trend of formulating semantic parsing as a machine translation problem. In particular, the sequence-to-sequence model \cite{sutskever2014sequence} is commonly used in recent works on semantic parsing \cite{dong-lapata-2018-coarse, jia-liang-2016-data, zhong-seq2sql}. Typically, they use a neural network encoder to encode the utterance sentence into a latent vector representation and use a decoder conditioned on the latent representation to predict the MRL as a sequence of symbols.

Due to the research interest in semantic parsing, many public datasets have been made available for English semantic parsing, ranging from small datasets that contain only a few hundred or a few thousand examples, such as GeoQuery \cite{zelle1996} and ATIS \cite{dahl-etal-1994-expanding}, to larger datasets with tens of thousands of question-logical form pairs, such as WikiSQL \cite{zhong-seq2sql} and Overnight \cite{wang-etal-2015-building}.

Multilingual semantic parsing, however, has only begun to draw research attention in more recent years. Therefore, very few datasets have been published for semantic parsing in non-English languages. So far, almost all of the multilingual semantic parsing datasets are manually translated from their English versions. Due to the cost of manual translation, they are limited to small datasets. For example, \citet{jones-etal-2012-semantic} translated the GeoQuery dataset into German, Greek, and Thai. \citet{susanto-lu-2017-neural} translated the ATIS dataset into Indonesian and Chinese. \citet{haas-riezler-2016-corpus} created the NLMaps dataset which contains around 2,400 queries to a geographic database in English. The authors translated the queries into German but kept the MRL annotation the same as that for English. Apart from the semantic parsing datasets for question answering, there are some multilingual datasets with other logical form representations, such as multilingual GraphQuestions with graphs as the meaning representation \cite{ReddyTPSL17}, Parallel Meaning Bank with DRT (Discourse Representation Theory) representation \cite{abzianidze-etal-2017-parallel}, and multilingual AMR test set with Abstract Meaning Representation \cite{damonte-cohen-2018-cross}. The logical form representation in these datasets are very different from the MRLs used for question answering and thus cannot be easily harnessed by many semantic parsers.

Among the limited literature on multilingual semantic parsing, several different methods have been proposed. The first attempts on multilingual semantic parsing \cite{haas-riezler-2016-corpus, damonte-cohen-2018-cross} use statistical/neural machine translation methods to translate non-English questions into English and rely on using an English semantic parser to parse all the utterances. Annotation projection is an alternative technique to deal with the lack of multilingual data. It maps the annotation from one language to another using word alignment. It has been applied to many NLP applications, including POS tagging \cite{yarowsky-etal-2001-inducing}, role-labeling \cite{akbik-etal-2015-generating}, semantic CCG parsing \cite{evang-bos-2016-cross}, and AMR parsing \cite{damonte-cohen-2018-cross}. In addition, \citet{susanto-lu-2017-neural} approached multilingual semantic parsing with a multi-task learning technique. They used separate encoders to encode sentences in different languages and used a shared decoder to predict the MRL. \citet{duong-etal-2017-multilingual} used cross-lingual word embeddings in a sequence-to-sequence model. They observed that using cross-lingual word embeddings improves the results on both English and German over their baseline models on the NLMaps dataset. They also compared training a model with a single encoder on multilingual data against training with separate encoders for each language and found that keeping separate encoders actually harms semantic parsing accuracy. Based on their observation, we will use a single encoder for multiple languages in our experiments. The role of machine translation to bootstrap crosslingual semantic parsers has also been investigated in \citet{sherborne-etal-2020-bootstrapping}.

\section{Multilingual Semantic Parsing Data}
\label{sec:data}

To tackle the data scarcity problem for multilingual semantic parsing, we aim to utilize machine translation to automatically generate training data from the more abundant English data for other languages. In this section, we introduce the English semantic parsing dataset we are using and describe our strategy to bootstrap training data for multilingual semantic parsing. 

\subsection{English Semantic Parsing Data}

We use the Facebook Task Oriented Parsing (TOP) dataset \cite{top-paper} as our source English semantic parsing data. The TOP dataset contains around 44k navigation and event questions created by crowd-sourced workers. The questions are annotated to semantic frames comprising of hierarchical intents and slots. We adapted the original intent-slot representation to a representation that is more similar to other question answering MRLs. More specifically, we dropped the text mentions in the intent label and kept only the entity text in the slot label. The resulting MRL is still a valid meaning representation because the text in the intent label does not affect the execution of the query on a knowledge base. Table \ref{tab:top_data} shows an example of the original TOP data and its corresponding MRL representation in the adapted task. 

We also remove the utterances where the root intent is \texttt{IN:UNSUPPORTED}, as it is a noisy catch-all class for out-of-domain utterances. The final dataset contains 28,414 training, 4,032 validation, and 8,241 test data points. 
\begin{table}[t!]
	\vspace{1em}
	\noindent\fbox{
		\parbox{\columnwidth}{
			\small
			\noindent  Question: \\
			\texttt{Any festivals this weekend} \\
			\\
			Hierarchical intent-slot representation: \\
			\texttt{{[}IN:GET\_EVENT Any \\   {[}SL:CATEGORY\_EVENT festivals {]} \\   {[}SL:DATE\_TIME this weekend {]}  {]} } \\
			\\
			Adapted MRL representation: \\
			\texttt{{[}IN:GET\_EVENT  \\   {[}SL:CATEGORY\_EVENT festivals {]} \\   {[}SL:DATE\_TIME this weekend {]}  {]} }
	}}
	
	\caption{An example of the English TOP dataset}
	\label{tab:top_data}
\end{table}

\subsection{Bootstrapping Multilingual Semantic Parsing Data}
Creating multilingual semantic parsing data from the English data is not a trivial task, because the MRL annotation is highly intertwined with the input question. Directly translating the text in the MRL into another language is likely to generate an incorrect MRL, as it may not match the translation of the input question. In order to obtain valid multilingual equivalents of both the natural language question and its meaning representation, rather than translating the MRL directly, we apply a similar method to annotation projection. We make use of the text alignment information between the question and the MRL to ensure that the translated MRL matches with the translated question. This is done in three steps:

\textbf{Step 1}: First, we reformat the question-MRL pair in English by replacing the text tokens in the MRL with placeholder tokens $x_0, x_1, ...$ that correspond to text tokens in the question. We also mark the positions of placeholder tokens in the question. Table \ref{tab:translation_step1} gives an example. 

\begin{table}[t]
	\vspace{1em}
	\noindent\fbox{
		\parbox{\columnwidth}{
			\small
			\noindent  Question (English): \\
			\texttt{Any festivals$|x_0$ this$|x_1$ weekend$|x_1$} \\
			\\
			MRL: \\
			\texttt{{[}IN:GET\_EVENT \\   {[}SL:CATEGORY\_EVENT $x_0$ {]}\\   {[}SL:DATE\_TIME  $x_1$ {]} {]} }
	}}
	
	\caption{Replacing text in the MRL with placeholder tokens and marking the positions of placeholder tokens in the question (on the same example as in Table \ref{tab:top_data}).}
	\label{tab:translation_step1}
\end{table}

\begin{figure}[h]
	\includegraphics[width=\columnwidth]{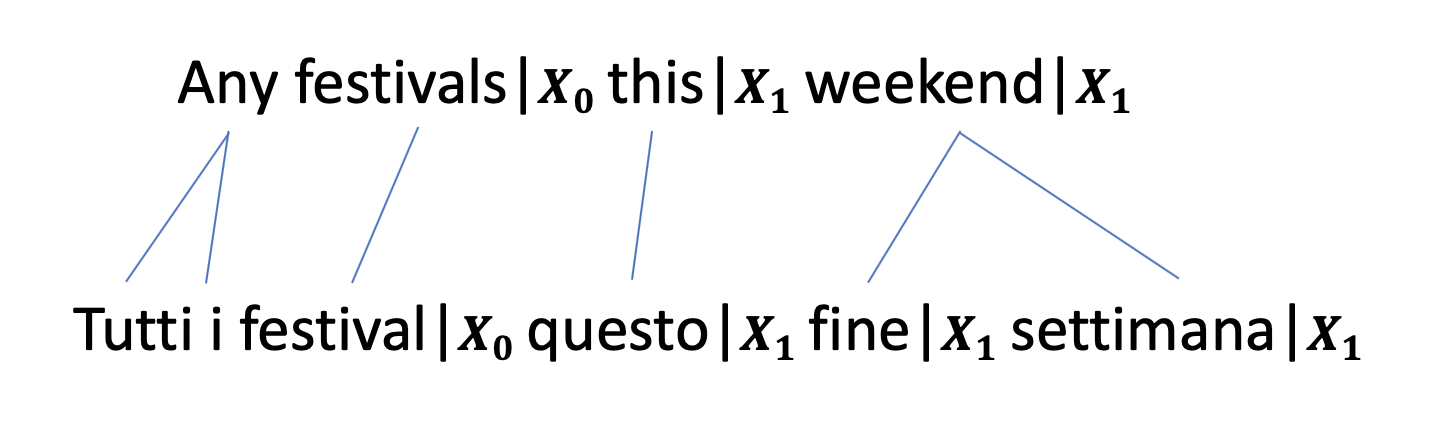}
	\caption{Using fast align algorithm to identify corresponding placeholder tokens in the translation.}
	\label{fig:alignment}
\end{figure}

\textbf{Step 2}: We then use the Amazon Machine Translation Service\footnote{\url{https://aws.amazon.com/translate/}} to translate the natural language question into the target language. Next, we use the fast align algorithm \cite{dyer-etal-2013-simple} to align the text between the translation and the original English sentence so as to identify the positions of the placeholder tokens in the translation. Figure \ref{fig:alignment} illustrates the alignment of texts between the source English sentence and its Italian translation and the identified placeholder tokens in the translation. 

\textbf{Step 3}: Finally, to obtain a valid MRL in the target language, we substitute the placeholder tokens in the MRL back with their corresponding text tokens in the translation. In this way, a valid pair of question and its MRL annotation in the target language is created (see Table \ref{tab:translation_step3}). 

\begin{table}[h]
	\vspace{1em}
	\noindent\fbox{
		\parbox{\columnwidth}{
			\small
			\noindent  Question (Italian): \\
			\texttt{Tutti i festival questo fine settimana} \\
			\\
			MRL: \\
			\texttt{{[}IN:GET\_EVENT \\   {[}SL:CATEGORY\_EVENT festival {]}\\   {[}SL:DATE\_TIME  questo fine settimana {]}  {]}}
	}}
	
	\caption{English semantic parsing data translated into Italian}
	\label{tab:translation_step3}
\end{table}

Following this method, we generate training data for Italian and Japanese semantic parsing from the English TOP dataset. We machine translated the training and validation splits of the TOP dataset into the two target languages. 

In order to evaluate the performance of our multilingual models on human-written sentences rather than machine-translated ones, we hire professional translators to manually translate the test set into Italian and Japanese. Table \ref{tab:top_multi_data} shows the data distribution of the multilingual TOP dataset. It should be noted that as the fast align algorithm may fail to align the tokens between the translation and the source text, especially when the source and target languages are dissimilar, we may lose some data points in the automatic multilingual data generation process. Overall, the vast majority of the training data can be bootstrapped successfully following our method (97.9\% data for Italian and 89.9\% for Japanese).

\begin{table}[ht]
	\centering
	\begin{tabular}{l|l|l|l}
		\hline
		Language & Train & Dev & Test \\
		\hline
		English & 28414 & 4032 & 8241\\
		Italian & 27830 & 3955 & 8241\\
		Japanese & 25544 & 3629 & 8241\\
		\hline
	\end{tabular}
	\caption{The distribution of the multilingual TOP dataset}
	\label{tab:top_multi_data}
\end{table}

\section{Multilingual Semantic Parsing Models}
\label{models}

\subsection{Model Architecture}
\label{sec:baseline}
Following the work of the state-of-the-art semantic parsers in English \cite{dong-lapata-2018-coarse, rongali2020don}, we base our multilingual semantic parsing model on the sequence-to-sequence method. We train a model that is similar in architecture to the Transformer encoder-decoder model described in Vaswani et al. \citep{transformer}. More specifically, we use a multilayer bidirectional Transformer encoder to encode the input question and a Transformer decoder to predict the MRL as a sequence of tokens. An encoder-decoder attention layer in the decoder learns to attend to the input tokens. We also implement an attention-based pointer mechanism \cite{NIPS2015_5866} to learn to copy text tokens from the input question. We concatenate the attention scores from the attention layer with the output vocabulary distribution from the final layer of the decoder. We then feed the concatenated vector to a Softmax layer to obtain a final probability distribution of possible actions. At each time step, the decoder either generates a symbol from the output vocabulary or outputs a pointer to one of the input tokens based on the scores from the final probability distribution. We use beam-search at inference time to select the prediction that maximizes the probability of the entire sequence. 
To train our baseline multilingual semantic parsing model, we mix the data from all languages together and train a single model from scratch to parse all questions. We apply Byte-Pair Encoding (BPE) tokenization \cite{sennrich-etal-2016-neural} to preprocess the data. BPE tokenization learns to break rare words into subword units. It is frequently used in machine translation and has contributed to better translation quality in many shared tasks \cite{denkowski-neubig-2017-stronger}. For multilingual tasks, we believe that subword representation helps to encode shared information between similar languages, and therefore facilitates multilingual semantic parsing. 

\subsection{Multilingual Semantic Parsing with Pretrained Encoders}
\label{sec:pretrain}
Transfer learning is a technique that aims to transfer information from a model trained on a source task to improve performance of the model on a target task. For neural network models, transfer learning typically consists of two stages: a pretraining stage and a finetuning stage. In the pretraining stage, the model is trained on the source task. In the finetuning stage, the knowledge of the trained model is transferred to the target task and adapted on that task. Existing literature has shown that transferring knowledge from pretrained models can improve the downstream performance on many NLP tasks \cite{devlin-etal-2019-bert}. 

As all our non-English semantic parsing training data are automatically generated from machine translation, it may not be as natural as real human-written sentences. We believe that transferring knowledge from a model that is pretrained on a huge amount of authentic multilingual text will allow our multilingual semantic parser to learn a better representation for the input utterance and to generalize better on real human-written sentences. To do that, we first initialize the encoder parameters with pretrained encoder parameters. We compare two state-of-the art multilingual encoders for initializing the multilingual semantic parser: the multilingual BERT (mBERT) model \cite{transformer} and the XLM-R model \cite{conneau2019cross}. Both models cover all the languages required in our semantic parsing tasks. The mBERT model is based on the multi-layer Transformer architecture. It is trained using the masked language objective and the next sentence prediction objective \cite{devlin-etal-2019-bert} on Wikipedia texts for the top 100 languages with the largest Wikipedia dumps. In our experiment, we use the public multilingual cased BERT-Base model\footnote{\url{https://github.com/google-research/bert/blob/master/multilingual.md}} (12-layer, 768-hidden, 12 heads) to initialize our semantic parsing encoder. The XLM-R model is a Transformer model trained using multilingual masked language model objectives \cite{conneau2019cross}. It is trained for 100 languages on the CommonCrawl corpus, which is several orders of magnitude larger than the Wikipedia dump, especially for low-resourced languages. We use the public XLM-R Base model\footnote{\url{https://github.com/pytorch/fairseq/tree/master/examples/xlmr}} (12-layer, 768-hidden, 12 heads) in our experiment. 

After initializing the semantic parsing model with pretrained encoder parameters, we finetune the models on the mixed multilingual semantic parsing data. To effectively adapt the pretrained encoder to our data, we implement gradual unfreezing \cite{howard-ruder-2018-universal} in the finetuning steps. Instead of tuning all encoder layers from the beginning, which may cause the model to forget what it learnt in pretraining, we slowly unfreeze the encoder layer weights to be tuned, from not changing the weights at all in the beginning until we finetune all the layers. 

\subsection{Baselines}
\label{sec:other_baselines}
We compare our multilingual semantic parsing models against two groups of baselines: monolingual models trained for each target language and a common method in previous research that also makes use of machine translation. 

\subsubsection{Monolingual Baselines}
We investigate how our multilingual semantic parsing models compare to monolingual models trained on each language separately. In accordance with the multilingual models, we build two types of monolingual baselines: monolingual models without pretraining and monolingual models finetuned from pretrained encoders. We use the same model architecture as the multilingual models for the monolingual baselines. For the monolingual pretrained encoders, we use the public English RoBERTa \cite{roberta} model to initialize the English model, because semantic parsers finetuned from the English RoBERTa model have achieved state-of-the-art result on the original TOP dataset \cite{rongali2020don}. As there is no public RoBERTa model available for Italian and Japanese, we use Italian and Japanese BERT models trained on Wikipedia data instead. 

In addition to using monolingual pretrained encoders, we also investigate a baseline with multilingual pretrained encoders (mBERT and XLM-R) finetuned on monolingual data for each target language.

\subsubsection{Multilingual Semantic Parsing through Machine Translation}

An alternative to multilingual semantic parsing is to translate all non-English languages into English and use an English semantic parsing model to understand the translated utterances \cite{haas-riezler-2016-corpus}. We compare our multilingual semantic parsing models against this method. We train an semantic parser on the English training data by finetuning from the RoBERTa model. We then use the Amazon Machine Translation Service to translate the Italian and Japanese sentences in the TOP test set into English. The translated texts are fed into the English semantic parser to get their MRL predictions. We use the MRL annotation of the English test set as the gold-standard for evaluation.

\begin{table*}[ht!]
	\centering
	\begin{tabular}{l|l|l|l}
		\hline
		Languages  & \begin{tabular}[c]{@{}l@{}}multilingual\\ (no pretraining)\end{tabular}   & mBERT & XLM-R  \\ \hline \hline
		English                                                        & 79.1\%      & 84.6\%          & \textbf{85.1\%}  \\ \hline
		Italian                                                          & 57.4\%      & 61.4\% &  \textbf{62.4\%}          \\ \hline
		Japanese                                                    & 31.9\%      & 34.2\%          & \textbf{36.3\%}  \\ \hline
		\hline
		Mixed                                                             & 56.1\%      & 60.1\%          & \textbf{61.2\%}      \\ \hline
	\end{tabular}
	\caption{Results on the multilingual TOP dataset}
	\label{tab:res_top}
\end{table*}

% \usepackage{multirow}
\begin{table*}[ht!]
	\begin{tabular}{l|l|l|l|l|l}
		\hline
		\multirow{2}{*}{Languages} & \multicolumn{4}{c|}{monolingual baselines} & \multirow{2}{*}{\begin{tabular}[c]{@{}l@{}}machine translated \\ to English\end{tabular}} \\ \cline{2-5}
		& no pretraining & monolingual BERTs & mBERT  & XLM-R  &                        \\ \hline \hline
		English  & 78.3\%         & 85.3\%            & 83.3\% & 83.8\% & 85.3\% (English model) \\ \hline
		Italian  & 55.9\%         & 55.1\%            & 59.8\% & 60.2\% & 35.3\%                 \\ \hline
		Japanese & 28.0\%         & 32.1\%            & 33.0\% & 32.5\% & 15.1\%                 \\ \hline
	\end{tabular}
	\caption{Results from baseline models on the multilingual TOP dataset}
	\label{tab:baselines}
\end{table*}

\section{Experiments and Results}
\label{sec:res}

\subsection{Experiment Setup}
We measure the performance of the semantic parsing models by exact match accuracy. By its definition, an MRL prediction is considered accurate only if the entire predicted sequence is exactly the same as the gold-standard MRL. The models are trained on AWS P3 instances with Tesla V100 GPU. We use the Adam optimizer in training and introduce early stopping if the loss doesn't improve on the validation set. We tune the hyperparameters for each model by random search on the validation set and report the results on the test set.

\subsection{Results and Analysis on the TOP Dataset}
\label{sec:top_res}
Table \ref{tab:res_top} shows the performance of the multilingual models on the TOP dataset and Table \ref{tab:baselines} shows the results of the baselines models. Comparing the multilingual models against the monolingual baselines, we find that training semantic parsing models on multilingual data jointly outperforms models trained on monolingual data only, even without using a pretrained encoder. The joint training is not only helpful for non-English languages, where the training data were machine translated, but it is also helpful for English, with or without a multilingual pretrained encoder.

In addition, we observe that transfer learning from pretrained encoders can improve the multilingual model performance further. Among the multilingual models, finetuning from pretrained XLM-R model achieves the best performance, which yields a parsing accuracy of 85.1\% for English, 62.4\% for Italian, and 36.3\% for Japanese. It substantially outperforms the monolingual baselines as well as the method that relies on machine translating utterances into English and using the English semantic parser to understand the utterances. The results prove that bootstrapping training data from English using machine translation is an effective method for constructing training data for multilingual semantic parsing.

On the other hand, constrained by the method we created our training data, the semantic parsing accuracy is heavily dependent on the machine translation quality. The better the machine translation model is, the more similar the automatically generated multilingual training data can be to real data. We measured the BLEU scores of the machine translation models on a random sample of English-Italian and English-Japanese sentences and found that the BLEU scores are 57.5 for Italian and 27.2 for Japanese, which shows that the English-Italian machine translation model is substantially more accurate than the English-Japanese one. Therefore, we observe a big difference between the semantic parsing accuracy for Italian and for Japanese. 

During error analysis, we find that a large group of errors in Italian semantic parsing is due to the inclusion or exclusion of articles copied in the MRL, which has minimal influence over the understanding. Table \ref{tab:missing_article} gives an example. As a heuristic solution, we filter out articles from both the expectation and the prediction and the exact match accuracy rises from 62.4\% to 75.4\% by our best performing model. Similarly, a large group of errors in Japanese is due to the inclusion or omission of postpositions and grammatical particles in the MRL when they are copied from the input question. If we filter out the postpositions and grammatical particles from the gold-standard and the predicted MRLs, the exact match accuracy is raised from 36.3\% to 52.3\%.

\begin{table}[ht!]
	\noindent\fbox{
		\parbox{\columnwidth}{
			\small
			\noindent  Question (Italian): \\
			\texttt{dove posso vedere i fuochi d'artificio questa sera} \\
			\\
			Gold-standard MRL: \\
			\texttt{[IN:GET\_EVENT [SL:CATEGORY\_EVENT i fuochi d'artificio ] [SL:DATE\_TIME questa sera ]]}
			\\
			
			Predicted MRL: \\
			\texttt{[IN:GET\_EVENT [SL:CATEGORY\_EVENT fuochi artificio ] [SL:DATE\_TIME questa sera ]]}
	}}
	
	\caption{An example of missing article ``i" in Italian semantic parsing}
	\label{tab:missing_article}
\end{table}

\subsection{Experiment on the NLMaps Dataset}
\label{sec:nlmaps}

% \usepackage{multirow}
\begin{table*}[]
	\centering
	\resizebox{\textwidth}{!}{
		\begin{tabular}{l|l|l|l|l|l|l|l}
			\hline
			\multirow{2}{*}{Languages} & \multicolumn{4}{c|}{monolingual baselines} & \multirow{2}{*}{\begin{tabular}[c]{@{}l@{}}multilingual\\ (no pretraining)\end{tabular}} & \multirow{2}{*}{mBERT} & \multirow{2}{*}{XLM-R} \\ \cline{2-5}
			& no pretraining & monolingual BERTs & mBERT & XLM-R &        &        &        \\ \hline \hline
			English & 73.5\%         & 74.1\%            & 75.7\%      &  63.3\%     & 72.1\% & \textbf{79.7\%} & 74.3\% \\ \hline
			German  & 68.0\%         & 70.3\%            & 71.6\%      & 59.5\%      & 66.9\% & \textbf{79.5\%} & 73.9\% \\ \hline \hline
			Mixed   & -              & -                 & -     & -     & 69.5\% & \textbf{79.6\%} & 74.1\% \\ \hline
	\end{tabular}}
	\caption{Results on the multilingual NLMaps dataset}
	\label{tab:res_nlmaps}
\end{table*}

\begin{table}[ht!]
	\centering
	\resizebox{\columnwidth}{!}{
		\begin{tabular}{l|c|c}
			\hline
			Languages &  Our best (mBERT) &  \citet{duong-etal-2017-multilingual}\\ \hline
			\hline
			English &  \textbf{85.9\%} &  85.7\%\\
			\hline
			German &  \textbf{85.5\%} &  82.3\%\\
			\hline
	\end{tabular}}
	\caption{Comparing the mBERT-based model with SOTA model on the NLMaps dataset (trained on the full training data for 10k iterations)}
	\label{tab:nlmaps_compare}
\end{table}

Apart from experimenting on the machine translated training data, we also want to see how our multilingual models perform with training data created by human and how our models compare to existing work. Therefore, we report the results of our models on the multilingual NLMaps dataset. The multilingual NLMaps dataset \cite{haas-riezler-2016-corpus} is one of the largest multilingual semantic parsing dataset published in previous literature. It contains around 2,400 English utterances and their manual translation into German. The queries are paired with a MRL representation that can be executed on a geographic database. Because NLMaps doesn't have a validation set, we randomly split 10\% of the training data as the validation set and trained our models on the remaining 90\% of the data. The resulting dataset contains 1,350 training utterance-MRL pairs, 150 validation pairs and 880 test pairs for both English and German.

Table \ref{tab:res_nlmaps} shows the results. Our best performing model on the NLMaps dataset is the multilingual semantic parser finetuned from the mBERT model, which yields an accuracy of 79.7\% for English and 79.5\% for German.  The best result reported on the multilingual NLMaps dataset in literature was by \citet{duong-etal-2017-multilingual}. However, their model was trained on the full training dataset for 10k iterations without splitting a separate validation set. Therefore, we retrain our best performing model under the same condition and present the result in Table \ref{tab:nlmaps_compare}. The result shows that our multilingual model outperforms the state-of-the-art model in German by 3.2\% while keeping the same level of accuracy in English (with a slight improvement of 0.2\%).

On the NLMaps dataset, we find that training the model on mixed multilingual data does not outperform a monolingual model if the models are trained without using a pretrained encoder. However, joint multilingual training is still helpful when a pretrained encoder is used. For example, when we finetune the mBERT model on English and German data separately, the resulting models yield an accuracy of 75.7\% for English and 71.6\% for German, which are markedly lower than the results from the mBERT model finetuned on mixed multilingual data. In addition, we find that using pretrained multilingual mBERT model outperforms pretrained monolingual BERT models.

\subsection{Experiment on Zero-shot Learning}

\label{sec:zero_shot}

\begin{table}[h!]
	\centering
	\resizebox{\columnwidth}{!}{
		\begin{tabular}{l|l|l|l}
			\hline
			\multicolumn{2}{l|}{Encoder weights unfreezing rate} & Italian         & Japanese       \\ \hline \hline
			\multirow{2}{*}{unfreeze all}          & mBERT        & 24.9\%          & 4.6\%          \\ \cline{2-4} 
			& XLM-R        & 36.1\%          & 1.7\%          \\ \hline
			\multirow{2}{*}{unfreeze 10\%}         & mBERT        & 28.6\%          & \textbf{7.3\%} \\ \cline{2-4} 
			& XLM-R        & \textbf{44.9\%} & 4.9\%          \\ \hline
			\multirow{2}{*}{freeze all}            & mBERT        & 16.0\%          & 2.5\%          \\ \cline{2-4} 
			& XLM-R        & 15.6\%          & 3.9\%          \\ \hline
	\end{tabular}}
	\caption{Zero-shot learning on the TOP dataset}
	\label{tab:zero-shot}
\end{table}

Zero-shot learning is a problem setup in which a model is tested on tasks that are not observed at training time. It studies the model's ability to generalize to unseen tasks. For multilingual models, we are interested in the zero-shot performance of a model when it is trained on one language and tested on other languages. To explore the zero-shot ability of our multilingual semantic parsers on the TOP dataset, we train a model with pretrained multilingual encoder on the English training data and apply the model to Italian and Japanese test data directly without further finetuning. We experiment with different ratios for unfreezing the pretrained encoder weights when tuning the models on the English data. Table \ref{tab:zero-shot} shows the results. We find that setting a small unfreezing rate to the pretrained encoder leads to a higher zero-shot accuracy. 

Multilingual models trained only on English data can achieve 44.9\% zero-shot accuracy when parsing Italian sentences, even though it has not seen any Italian semantic parsing data in training. Table \ref{tab:zero_shot_examples} shows an example. However, their zero-shot performance on Japanese sentences is very poor. This is not surprising as English and Italian are more similar and they share a lot more BPE subword units than English and Japanese. 

\begin{table}[t]
	\noindent\fbox{
		\parbox{\columnwidth}{
			\small
			\noindent  Question (Italian): \\
			\texttt{Concerti di Beyonce questo fine settimana} \\
			\\
			Predicted MRL: \\
			\texttt{[IN:GET\_EVENT [SL:CATEGORY\_EVENT Concerti ] [SL:NAME\_EVENT Beyonce ] [SL:DATE\_TIME questo fine settimana ] ]}
	}}
	
	\caption{An example of correct zero-shot prediction}
	\label{tab:zero_shot_examples}
\end{table}

\section{Conclusion}

In this paper, we describe our method to build multilingual semantic parsing models when the multilingual data is limited. We introduce a new multilingual semantic parsing dataset in English, Italian and Japanese based on the public TOP dataset, with training and validation data automatically generated from English and 8k test data manually translated. The multilingual TOP test set is so far the largest dataset for multilingual semantic parsing, which will be useful for future research. By leveraging joint multilingual training and transfer learning from pretrained encoders, our semantic parsing models outperform several baselines on the TOP dataset and the state-of-the-art on the NLMaps dataset. We show that semantic parsing models with pretrained multilingual encoders can generalize from English to Italian with 44.9\% zero-shot accuracy. However, we find that there is a gap between Italian and Japanese semantic parsing with our method. In future work, we plan to improve our models with both language-invariant and language-specific encodings and apply our method to more languages.

\bibliography{anthology,starsem2021}
\bibliographystyle{acl_natbib}

\end{document}